\documentclass{article}

\usepackage{natbib}

\usepackage[final]{ewrl_2024}

\usepackage[utf8]{inputenc} 
\usepackage[T1]{fontenc}    
\usepackage{hyperref}       
\usepackage{url}            
\usepackage{booktabs}       
\usepackage{amsfonts}       
\usepackage{nicefrac}       
\usepackage{microtype}      
\usepackage{xcolor}         

\usepackage{hyperref}
\usepackage{url}

\usepackage{microtype}
\usepackage{graphicx}
\usepackage{booktabs}

\usepackage{algorithm}
\usepackage[noend]{algpseudocode}
\usepackage{caption}
\usepackage{subcaption}
\usepackage{float}
\usepackage{amsmath}
\usepackage{amssymb}
\usepackage{mathtools}
\usepackage{amsthm}
\usepackage{bm}
\usepackage{wrapfig}

\usepackage{tikz}
\usepackage{tabularx}
\usepackage{makecell}
\usepackage{float}
\usepackage{enumitem}
\setlist{nolistsep}
\usepackage{multirow}

\theoremstyle{plain}

\newtheorem{proposition}{Proposition}

\newtheorem{appproposition}{Proposition}

\theoremstyle{definition}
\newtheorem{theoretical_result}{Theoretical Result}
\newtheorem{hypothesis}{Hypothesis}

\usepackage{mathbbol}
\DeclareSymbolFontAlphabet{\amsmathbb}{AMSb}

\usepackage[most]{tcolorbox}
\tcbuselibrary{skins}
\newtcolorbox{gbox}{
    width=0.9\textwidth,
    center,
    interior hidden,
    colback=white, 
    boxsep=3pt,
    left=1pt,
    right=1pt,
    top=1pt,
    bottom=1pt
}

\definecolor{red}{RGB}{214,39,40}
\definecolor{newblue}{RGB}{31,119,180}

\newcommand{\theor}{Theoretical Result }
\newcommand{\hypo}{Hypothesis }
\newcommand{\theorlowp}{theoretical results }
\newcommand{\hypolowp}{hypotheses }
\newcommand{\sgenv}{Simple Gridworld }
\newcommand{\dt}{decision-time }
\newcommand{\bg}{background }
\newcommand{\rrl}{regular RL }
\newcommand{\tl}{transfer learning }
\newcommand{\simple}{simplest instantiations }
\newcommand{\modern}{modern instantiations }

\newcommand{\algfontsize}{\footnotesize }

\title{A Look at Value-Based Decision-Time vs.\ Background Planning Methods Across Different Settings}

\author{
  Safa Alver \\
  Mila, McGill University \\
  Montreal, QC, Canada \\
  \texttt{safa.alver@mail.mcgill.ca} \\
  \And
  Doina Precup \\
  Mila, McGill University and Google DeepMind \\
  Montreal, QC, Canada \\
  \texttt{dprecup@cs.mcgill.ca} \\
}

\begin{document}

\maketitle

\begin{abstract}
  In model-based reinforcement learning (RL), an agent can leverage a learned model to improve its way of behaving in different ways. Two of the prevalent ways to do this are through decision-time and background planning methods. In this study, we are interested in understanding how the value-based versions of these two planning methods will compare against each other across different settings. Towards this goal, we first consider the simplest instantiations of value-based decision-time and background planning methods and provide theoretical results on which one will perform better in the regular RL and transfer learning settings. Then, we consider the modern instantiations of them and provide hypotheses on which one will perform better in the same settings. Finally, we perform illustrative experiments to validate these theoretical results and hypotheses. Overall, our findings suggest that even though value-based versions of the two planning methods perform on par in their simplest instantiations, the modern instantiations of value-based decision-time planning methods can perform on par or better than the modern instantiations of value-based background planning methods in both the regular RL and transfer learning settings.
\end{abstract}

\section{Introduction}
\label{sec:intro}

It has long been argued that, models will play a key role in building truly intelligent agents that can perform counterfactual reasoning and fast re-planning \citep{sutton2018reinforcement, sutton2022alberta, schaul2018barbados}. Although this is a generally-agreed view in the reinforcement learning (RL) community, the question of how to leverage a model to perform planning does not have a widely-accepted answer. In model-based RL, two of the prevalent ways to perform planning are through \dt and \bg planning methods, where the agent mainly plans in the moment and in parallel to its interaction with the environment, respectively \citep{sutton2018reinforcement}. Even though these two planning methods have been developed with different scenarios and application domains in mind: (i) \dt planning algorithms \citep{tesauro1996line, silver2017mastering, silver2018general} for scenarios in which the exact model of the environment is known and for domains that allow for an adequate computational budget at decision time such as board games like chess and Go, and (ii) \bg planning algorithms \citep{sutton1990integrated, sutton1991dyna, kaiser2020model, hafner2021mastering, hafner2023mastering} for scenarios in which the exact model is to be learned through pure interaction with the environment and for domains that are agnostic to the response time of the agent such as gridworlds and video games, recently, with the introduction of the ability to learn a model through pure interaction, \dt planning algorithms have been applied to the same scenarios and domains as their \bg planning counterparts: see e.g. \citet{schrittwieser2020mastering} and \citet{hamrick2021role} which both evaluate a \dt planning algorithm, called MuZero, on the Atari suite \citep{bellemare2013arcade}. However, it still remains unclear in which settings will one of these planning methods perform better than the other.

In this study, we attempt to provide an answer to a specific version this question. Specifically, we are interested in the following question:

\begin{gbox}
    \textit{Using the discounted return as the performance measure and considering value-based versions of these methods, in which settings will one planning method perform better than the other?} 
\end{gbox}

To answer this question, we first start by considering the simplest instantiations of value-based decision-time and background planning methods and provide \theorlowp on which one will perform better in the \rrl and \tl settings. Then, we consider the modern instantiations of them and provide hypotheses on which one will perform better in the same settings. Finally, we perform illustrative experiments to validate these \theorlowp and hypotheses. Overall, our results suggest that even though value-based \dt and \bg planning methods perform on par in their simplest instantiations, due to (i) not using simulated experience in the updates of the value estimator, and (ii) the ability to improve upon the existing policy in generalization settings, the \modern of value-based \dt planning methods can perform on par or better than the \modern of value-based \bg planning methods in both the \rrl and \tl settings. We hope that our findings will help the RL community towards developing a better understanding of how the two value-based planning methods will compare against each other across different setting, and thereby stimulate research in improving them in potentially interesting ways.

\textbf{Key Contributions.} The key contributions of this study are as follows: (i) We provide a unified analysis of the value-based versions of the two planning methods and explain how planning is performed within each of these methods (Sec.\ \ref{sec:background}). (ii) We provide theoretical results and hypotheses on how the value-based versions of the two planning methods will compare against each other across different settings (Sec.\ \ref{sec:DTvsB_planning}), and (iii) we provide illustrative empirical results that validate these theoretical results and hypotheses (Sec.\ \ref{sec:experiments}).

\section{Background}
\label{sec:background}

\textbf{Reinforcement Learning.} In value-based RL \citep{sutton2018reinforcement}, an agent interacts with its environment through a sequence of actions to maximize its long-term cumulative reward. Here, the environment is usually modeled as a Markov decision process $(\mathcal{S}, \mathcal{A}, p, r, \rho, \gamma)$, where $\mathcal{S}$ and $\mathcal{A}$ are the (finite) set of states and actions, $p:\mathcal{S}\times \mathcal{A} \to \text{Dist}(\mathcal{S})$ is the transition distribution, $r:\mathcal{S}\times \mathcal{A}\times \mathcal{S}\to \amsmathbb{R}$ is the reward function, $\rho:\mathcal{S}\to \text{Dist}(\mathcal{S})$ is the initial state distribution, and $\gamma\in [0,1)$ is the discount factor. The goal of the agent is to learn, through pure interaction, a value estimator $Q:\mathcal{S}\times \mathcal{A}\to \amsmathbb{R}$ that induces a policy $\pi \in \mathbb{\Pi} \equiv \{\pi | \pi:\mathcal{S}\to \text{Dist}(\mathcal{A}) \}$, maximizing $E_{\pi, p} [\sum_{t=0}^\infty \gamma^t r(S_t, A_t, S_{t+1}) | S_0 = s_0 \sim \rho]$.

\textbf{Model-Based RL.} One way of achieving this goal is through the use of value-based model-based RL methods, in which there are two alternating phases: the learning and planning phases.\footnote{We note that even though some model-based RL algorithms, such as AlphaGo/AlphaZero \citep{silver2017mastering, silver2018general}, do not employ a model learning phase and make use of a priori given exact model, in this study, we only study algorithms in which the model has to be learned from pure interaction with the environment.} In the learning phase, the gathered experience is mainly used in learning a model $m \in \mathcal{M} \equiv \{(p,r,\rho) | p:\mathcal{S}\times\mathcal{A}\to \text{Dist}(\mathcal{S}), r:\mathcal{S}\times\mathcal{A}\times\mathcal{S}\to \amsmathbb{R}, \rho:\mathcal{S}\to \text{Dist}(\mathcal{S}) \}$, and it may also be used in improving the value estimator. In the planning phase, the learned model $m$ is then used for simulating experience, either to be used alongside real experience in improving the value estimator or just to be used in selecting actions at decision time. In value-based model-based RL methods, similar to value-based RL methods, the agent only employs a value estimator and a model, i.e.\ it does not employ an explicit policy.

\textbf{Value-Based Planning Methods.} In value-based model-based RL, two of the prevalent ways to perform planning are through value-based \dt and \bg planning methods \citep[see Ch.\ 8 of][]{sutton2018reinforcement}.\footnote{Although some new planning methods have been proposed in the transfer learning literature (see e.g., \citet{barreto2017successor, barreto2019option, barreto2020fast, alver2022constructing}), these approaches can also be viewed as performing some form of \dt planning with pre-learned models.} For the sake of presentation clarity, in the rest of this study, we will drop the ``value-based'' adjective and refer to these planning methods as just \dt and \bg planning methods. In decision-time planning methods, planning is performed as a computation whose output is the selection of a single action for the current state, which is done by unrolling the model forward from the current state to compute local value estimates. Here, planning is performed independently for every encountered state and it is mainly performed in an online fashion, though it may also contain offline components. In contrast, in \bg planning methods, planning is performed by continually improving a value estimator, on the basis of simulated experience from the model, in a global manner. Action selection is then quickly done by querying the value estimator at the current state. Unlike \dt planning, \bg planning is performed in a purely offline fashion, in parallel to the agent-environment interaction, and thus is not focused on the current state. For convenience, in this study, we will refer to all model-based RL algorithms that have an online planning component as decision-time planning algorithms \citep[see e.g.][]{tesauro1996line, silver2017mastering, silver2018general, zhao2021consciousness}, and will refer to the rest as background planning algorithms \citep[see e.g.][]{sutton1990integrated, sutton1991dyna, kaiser2020model, zhao2021consciousness}.

It is important to note that in decision-time planning methods, the simulated experience is \textit{only} used for selecting actions at decision time and it is \textit{not} used for updating the value estimator; the value estimator is updated with only \textit{real} experience. In contrast, in background planning methods, the simulated experience is used \textit{alongside} the real experience for updating the value estimator.

\begin{wrapfigure}{R}{0.57\textwidth}
    \centering
    \begin{subfigure}{0.23\textwidth}
        \centering
        \includegraphics[height=2.5cm]{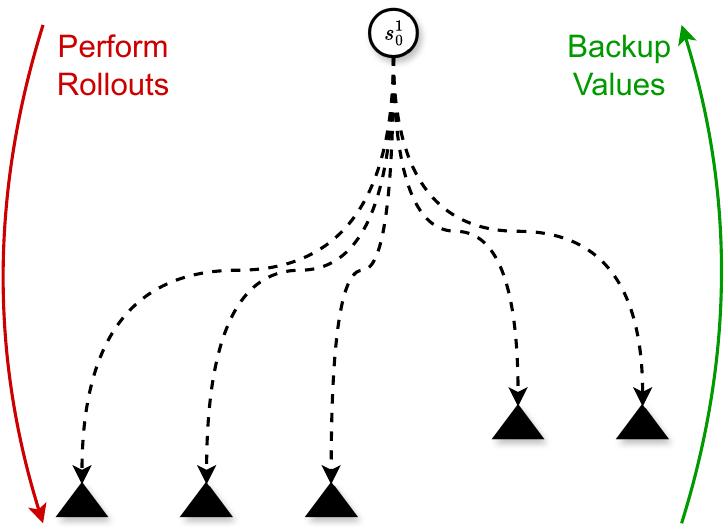}
        \caption{Pure Rollouts} \label{fig:pure_rollout}
    \end{subfigure}
    \begin{subfigure}{0.32\textwidth}
        \centering
        \includegraphics[height=2.5cm]{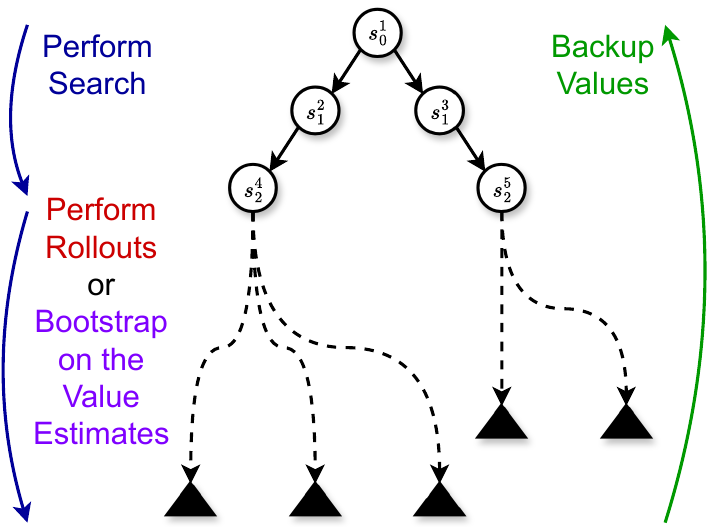}
        \caption{Search + Rollouts/Bootstrap} \label{fig:search_and_rollout}
    \end{subfigure}
    \caption{The different planning methods within \dt planning in which planning is done (a) by purely performing rollouts, and (b) by first performing some amount of search and then by either performing rollouts or bootstrapping on value estimates. The subscripts and superscripts on the states indicate the time steps and state identifiers, respectively. The black triangles indicate the terminal states.}
    \vspace{-0.5cm}
    \label{fig:dt_planning_figures}
\end{wrapfigure}

\textbf{Categorization within the Two Planning Methods.} Starting with \dt planning, depending on how much search is performed, \dt planning methods can be studied under two main categories: 
\begin{enumerate}
    \item Decision-time planning methods that do not perform search
    \item Decision-time planning methods that perform search
\end{enumerate}
In the first category, planning is done by just running rollouts (see Fig.\ \ref{fig:pure_rollout}). A notable example of a planning method in this category is the online Monte-Carlo planning algorithm of \citet{tesauro1996line}. In the second one, planning is done by first performing some amount of search and then either (i) by running rollouts or (ii) by bootstrapping on the value estimates (see Fig.\ \ref{fig:search_and_rollout}). A notable example of a planning method in this category is the deep model-predictive control algorithm of \citet{zhao2021consciousness}. We refer the reader to \citet{bertsekas2021rollout} for more details on the different categories within \dt planning methods.

Moving on to \bg planning, as all \bg planning methods perform planning by continually updating a value estimator with simulated experience generated by the model, we do not study them under different categories. Notable examples of background planning methods include the Dyna-Q algorithm of \citet{sutton1990integrated, sutton1991dyna} and the deep Dyna-Q algorithm of \citet{zhao2021consciousness}.

\textbf{Value-Equivalent Models.} One of the recent trends in model-based RL is to learn models that are specifically useful for planning \citep{grimm2020value, grimm2021proper}. Following \citet{grimm2021proper}, we define value-equivalent models as follows. Let $V_{m}^{\pi}\in\amsmathbb{R}^{|\mathcal{S}|}$ be the value vector of a policy $\pi\in\mathbb{\Pi}$ evaluated in model $m\in\mathcal{M}$, whose elements are defined $\forall s\in\mathcal{S}$ as $V_{m}^{\pi}(s) = E_{\pi, p} \left[ \sum_{t=0}^\infty \gamma^t r(S_t, A_t, S_{t+1}) \vert S_0 = s \right]$, and let $V_{m}^*\in\amsmathbb{R}^{|\mathcal{S}|}$ be the optimal value vector in model $m$. We say that a model $m_{\text{VE}}\in\mathcal{M}$ is a \emph{value-equivalent (VE) model} of the true environment $m^*\in\mathcal{M}$ if the following equality holds:
\begin{equation*}
    V_{m^*}^{\pi_{m_{\text{VE}}}^*} = V_{m^*}^* \quad \forall \pi_{m_{\text{VE}}}^*\in\mathbb{\Pi},
\end{equation*}
where $\pi_{m_{\text{VE}}}^*$ is an optimal policy obtained as a result of planning with model $m_{\text{VE}}$.

\textbf{Evaluation Settings in RL.} In RL, two of the most commonly-used evaluation settings are the regular RL and transfer learning settings. In the regular RL setting, the agent is trained and evaluated on the same task (or distribution of tasks). Although there are many different subcategories within the \tl setting \citep{taylor2009transfer, zhu2023transfer}, in this study we will only consider the adaptation and generalization settings as these settings are commonly-used for evaluating the simplest and modern instantiations of the two planning styles. In adaptation settings, the agent is first trained and evaluated on a source task (or a distribution of source tasks) and then it is trained and evaluated on a subsequent target task (or a distribution of target tasks). In generalization settings, the agent is just evaluated on a target task (or a distribution of target tasks) as it gets trained and evaluated on a source task (or a distribution of source tasks).

\section{Decision-Time vs.\ Background Planning Methods Across Different Settings}
\label{sec:DTvsB_planning}

In this study, we are interested in understanding in which settings will one planning method perform better than the other. Thus, we start by formally defining a performance measure that will be used in comparing the two planning methods of interest. Given an arbitrary model $m=(p,r,\rho)\in\mathcal{M}$, we define the performance of an arbitrary policy $\pi\in\mathbb{\Pi}$ in it as follows: $J_{m}^{\pi} = E_{\pi, p} [ \sum_{t=0}^{\infty} \gamma^t r(S_t, A_t, S_{t+1}) | S_0 = s_0 \sim \rho ]$. We note that $J_{m}^{\pi}$ corresponds to the expected discounted return of a policy $\pi$ in model $m$.

\subsection{Simplest Instantiations with Tabular Representations}
\label{sec:classic_inst}

\begin{wrapfigure}{R}{0.65\textwidth}
\vspace{-0.75cm}
\centering
\begin{minipage}{1.0\linewidth}
    \begin{algorithm}[H]
    \footnotesize
    \centering
    \caption{The joint pseudocode of the OMCP algorithm (with an adaptable model and improving rollout policy) 
    and the Dyna-Q algorithm. The \textcolor{newblue}{blue} and \textcolor{red}{red} colored parts are specific to the \textcolor{newblue}{OMCP} and \textcolor{red}{Dyna-Q} algorithms, respectively. That is, OMCP does not contain the red parts and Dyna-Q does not contain the blue parts.}\label{alg:alg_SI}
    \begin{algorithmic}[1]
        \State \text{Initialize $Q$ and $m$ $\forall s\in\mathcal{S}$ and $\forall a\in\mathcal{A}$}
        \State \textcolor{newblue}{$n_r\gets \text{number of episodes to perform rollouts}$}
        \State \textcolor{red}{$n_p\gets \text{number of iterations to perform planning}$}
        \For{each episode}
        \State $S\gets \text{reset environment}$
        \While{\text{not done}}
        \State $A\gets$ \textcolor{newblue}{$\epsilon\text{-greedy}(\text{rollouts}(S, m, n_r, Q))$} / \textcolor{red}{$\epsilon\text{-greedy}(Q(S,\cdot))$}
        \State \text{$R, S', \text{done} \gets \text{environment($A$)}$}
        \State \text{Update} $Q(S,A)$ and $m(S,A)$ \text{with $R$, $S'$, $\text{done}$}
        \For{\textcolor{red}{$i\in \{1,\dots, n_p\}$}}
        \State \textcolor{red}{$S, A \gets \text{sample from } \mathcal{S}\times\mathcal{A}$}
        \State \textcolor{red}{$R,S', \text{done}\gets m(S,A)$}
        \State \textcolor{red}{\text{Update} $Q(S,A)$ \text{with} $R$, $S'$, $\text{done}$}
        \EndFor
        \State $S\gets S'$
        \EndWhile
        \EndFor
    \end{algorithmic}
\end{algorithm}
\end{minipage}
\vspace{-0.25cm}
\end{wrapfigure}

For easy analysis, we start by considering the simplest instantiations of the two planning methods with tabular value estimators and models, which can be found in Ch.\ 8 of \citet{sutton2018reinforcement}. More specifically, for \dt planning we study the online Monte-Carlo planning (OMCP) algorithm of \citet{tesauro1996line}, which falls under the category of decision-time planning algorithms that do not perform search (see Sec.\ \ref{sec:background}), and for \bg planning we study the Dyna-Q algorithm of \citet{sutton1990integrated, sutton1991dyna}. The pseudocode that shows the similarities and differences between these two algorithms is presented in Alg.\ \ref{alg:alg_SI}; the only difference between the two algorithms is in how they perform planning (see line 7 for OMCP and lines 10-13 for Dyna-Q), which allows for performing controlled experiments.

\textbf{Regular RL Setting.} We start by considering the regular RL setting in which an agent is trained and evaluated on the same task. In this setting, the following statement can easily be proven:

\begin{proposition} 
    \label{prop:prop}
    Let $m_{O}\in\mathcal{M}$ and $m_{D}\in\mathcal{M}$ denote the models of the of the OMCP and Dyna-Q algorithms, respectively, and let $\pi_{m_O}$ and $\pi_{m_D}$ denote their final policies generated as a result of planning with their corresponding models. Then, when $m_O$ and $m_D$ become VE models of $m^*\in\mathcal{M}$, we have $J_{m^*}^{\pi_{m_O}} = J_{m^*}^{\pi_{m_D}}$.
\end{proposition}

Due to space constraints, we defer the proof to App.\ \ref{app_sec:proofs}. Prop.\ \ref{prop:prop} states that, the OMCP and Dyna-Q algorithms will perform on par when their models become VE models. More explicitly, in the regular RL setting, we expect the following statement to hold:

\begin{gbox}
    \begin{theoretical_result}
    \label{tr:tr1}
        In the regular RL setting, the OMCP algorithm will perform on par with the Dyna-Q algorithm.
    \end{theoretical_result}
\end{gbox}

\textbf{Transfer Learning Setting.} In the transfer learning setting, we consider an adaptation setting that is commonly-used for evaluating the simplest instantiations of the two planning methods \citep[see e.g.\ Ch.\ 8 of][]{sutton2018reinforcement}. In this setting, the agent is first trained and evaluated on a source task and then it is trained and evaluated on a subsequent target task. In this setting, Prop.\ \ref{prop:prop} of the \rrl setting will hold directly, as the adaptation setting consists of two consecutive \rrl settings, one with the source task and the other with the subsequent target task. Restating more clearly, in the adaptation setting, we expect the following statement to hold:

\begin{gbox}
    \begin{theoretical_result}
    \label{tr:tr2}
        In the adaptation setting, the OMCP algorithm will perform on par with the Dyna-Q algorithm in the source task, and the same will happen in the subsequent target task.
    \end{theoretical_result}
\end{gbox}

We note that, although we only considered the OMCP and Dyna-Q algorithms, the statements that we provide in this section are generally applicable to most other \simple of the two value-based planning methods \citep{sutton2018reinforcement} as these instantiations also perform planning \& learning in a similar fashion described in Alg.\ \ref{alg:alg_SI} and they also use tabular representations in their value estimators and models.

\subsection{Modern Instantiations with Neural Network Representations}
\label{sec:modern_inst}

\begin{wrapfigure}{R}{0.6\textwidth}
\vspace{-0.75cm}
\centering
\begin{minipage}{1.0\linewidth}
    \begin{algorithm}[H]
    \footnotesize
    \centering
    \caption{The joint pseudocode of the deep MPC and deep Dyna-Q algorithms. The \textcolor{newblue}{blue} and \textcolor{red}{red} colored parts are specific to the \textcolor{newblue}{deep MPC} and \textcolor{red}{deep Dyna-Q} algorithms, respectively. That is, deep MPC does not contain the red parts and deep Dyna-Q does not contain the blue parts. Here, the imaginary replay buffer stores the state-action pairs that are to be used in generating simulated experience.}\label{alg:alg_MI}
    \begin{algorithmic}[1]
        \State \text{Initialize the parameters $\theta$ and $\omega$ of $Q_{\theta}$ and $m_{\omega}$}
        \State Initialize the replay buffer $\mathcal{D}\gets \{ \}$
        \State \textcolor{red}{\text{Initialize the imaginary replay buffer $\mathcal{D}_i\gets \{ \}$}}
        \State $N\gets \text{replay buffer size to start performing updates}$
        \State \textcolor{newblue}{$n_s\gets \text{number of steps to perform search}$}
        \State \textcolor{red}{$n_{ib}\gets \text{number of samples to sample from } \mathcal{D}_i$}
        \State $n_{b}\gets \text{number of samples to sample from } \mathcal{D}$
        \State \textcolor{newblue}{$h\gets \text{search heuristic}$}
        \For{each episode}
        \State $S\gets \text{reset environment}$
        \While{\text{not done}}
        \State $A\gets$ \textcolor{newblue}{$\epsilon\text{-greedy}(\text{search+bootstrap}(S, m_\omega, n_s, h, Q_{\theta}))$} / \textcolor{red}{$\epsilon\text{-greedy}(Q_{\theta}(S,\cdot))$}
        \State \text{$R, S', \text{done} \gets \text{environment($A$)}$}
        \State \text{$\mathcal{D}\gets \mathcal{D} + \{(S,A,R,S', \text{done})\} $}
        \State \textcolor{red}{\text{$\mathcal{D}_i\gets \mathcal{D}_i + \{(S,A)\} $}}
        \If{$|\mathcal{D}| \geq N$}
        \State \textcolor{red}{$\mathcal{B}_i\gets \text{samplebatch}(\mathcal{D}_i, n_{ib})$}
        \State \textcolor{red}{$\mathcal{B}_i\gets \mathcal{B}_i + m_\omega(\mathcal{B}_i)$}
        \State $\mathcal{B}\gets \text{samplebatch}(\mathcal{D}, n_{b})$
        \State Update $Q_{\theta}$ with \textcolor{newblue}{$\mathcal{B}$} / \textcolor{red}{$\mathcal{B}_i + \mathcal{B}$}
        \State \text{Update} $m_\omega$ \text{with $\mathcal{B}$}
        \EndIf
        \State $S\gets S'$
        \EndWhile
        \EndFor
    \end{algorithmic}
\end{algorithm}
\end{minipage}
\vspace{-0.5cm}
\end{wrapfigure}

We now consider the \modern of the two planning methods which represent their models with neural networks. More specifically, for \dt planning we study the deep model-predictive control (MPC) algorithm of \citet{zhao2021consciousness}, which falls under the category of decision-time planning algorithms that perform search (see Sec.\ \ref{sec:background}), and for \bg planning we study the deep Dyna-Q algorithm of \citet{zhao2021consciousness}. The pseudocode that shows the similarities and differences between these algorithms is presented in Alg.\ \ref{alg:alg_MI}. We choose to study these algorithms as the only difference between them is in how they perform planning (see line 12 for deep MPC and lines 17-20 for deep Dyna-Q), that is, similar to the algorithms in Sec.\ \ref{sec:classic_inst}, these algorithms share the same architectures and learning backbones and only differ in their planning methods, which allows for performing controlled experiments.

\textbf{Regular RL setting.} In the regular RL setting, as in the case with the OMCP and Dyna-Q algorithms, we would normally expect  the deep MPC and deep Dyna-Q algorithms to perform similarly. However, as neural networks are used in the representations of the value estimators and models, the deep Dyna-Q algorithm can suffer from updating its value estimator with the harmful simulated experiences generated by its pre-VE models, which are earlier versions of the model before it becomes a VE model (see the red part in line 20 of Alg.\ \ref{alg:alg_MI}). More specifically, as neural networks are used in the representation of the value estimator, the harmful simulated experiences from the pre-VE models can cause permanent damages in the learning dynamics of the value estimator and slowdown, or prevent, the deep Dyna-Q algorithm in reaching optimal (or good) performance
\citep[see e.g.][]{van2019use, jafferjee2020hallucinating}.
Even though, the pre-VE models of the deep MPC algorithm will also generate harmful simulated experiences, this will have no permanent effects on its performance, as in deep MPC the value estimator is only updated with real experiences coming from the environment (see the blue part in line 20 of Alg.\ \ref{alg:alg_MI}). Thus, we hypothesize that the deep MPC algorithm will perform on par or better than the deep Dyna-Q algorithm, or more explicitly:

\begin{gbox}
    \begin{hypothesis}
    \label{h:h1}
        In the regular RL setting, as the value estimator of the deep Dyna-Q algorithm will be suffering from the negative effects of the harmful simulated experiences, the deep MPC algorithm will perform on par or better than the deep Dyna-Q algorithm.
    \end{hypothesis}
\end{gbox}

\textbf{Transfer Learning Setting.} In the transfer learning setting, we consider adaptation and generalization settings that are commonly-used for evaluating the \modern of the two planning methods \citep[see e.g.][respectively]{van2020loca, anand2022procedural}. In the adaptation setting, the agent is first trained and evaluated on a distribution of source tasks and then it is trained and evaluated on a distribution of subsequent target tasks. In the generalization setting, the agent is just evaluated on a distribution of target tasks as it gets trained and evaluated on a distribution of source tasks.

In both settings, harmful simulated experiences can again prevent the deep Dyna-Q algorithm from reaching optimal or good performance on the source tasks because of the same reasons discussed in the \rrl setting. Additionally, in the adaptation setting, after the tasks switch from the source tasks to the target tasks, deep Dyna-Q can suffer more in the adaptation process, as its model will keep generating experiences that resembles the source tasks until it adapts to the target tasks, which in the meantime will lead harmful updates to the value estimator. Also, in the generalization setting, if the learned model of the deep MPC algorithm becomes capable of simulating at least a few time steps of the target tasks and if the learned policies of both algorithms perform similarly on the target tasks, the deep MPC algorithm will perform better on the target tasks, as while being evaluated on the target tasks it will be able to improve upon its existing policy via performing online planning; improving upon its existing policy is not possible for the deep Dyna-Q algorithm as it performs planning in an offline fashion and thus requires additional interaction with the target tasks, which is not possible in the generalization settings. More concretely, in the adaptation and generalization settings, we hypothesize that:

\begin{gbox}
    \begin{hypothesis}
    \label{h:h2}
        In both the adaptation and generalization settings, a similar performance trend with \hypo \ref{h:h1} will hold on the source tasks.
    \end{hypothesis}
    \begin{hypothesis}
    \label{h:h3}
        In the adaptation setting, the deep Dyna-Q algorithm will suffer more in the adaptation process and thus perform worse than the deep MPC algorithm on the target tasks.
    \end{hypothesis}
    \begin{hypothesis}
    \label{h:h4}
        In the generalization setting, the deep MPC algorithm will improve upon its existing policy and perform better than the deep Dyna-Q algorithm on the target tasks.
    \end{hypothesis}
\end{gbox}

We note again that although we only considered the deep MPC and deep Dyna-Q algorithms, the statements that we provide in this section are also generally applicable to most other modern instantiations of the two value-based planning methods \citep{moerland2023model} as they also perform planning \& learning in a similar fashion described in Alg.\ \ref{alg:alg_MI} and also represent their value estimators and models with neural networks.

\section{Experimental Results}
\label{sec:experiments}

\begin{wrapfigure}{R}{0.58\textwidth}
    \vspace{-0.25cm}
    \centering
    \begin{subfigure}{0.14\textwidth}
        \centering
        \includegraphics[height=1.5cm]{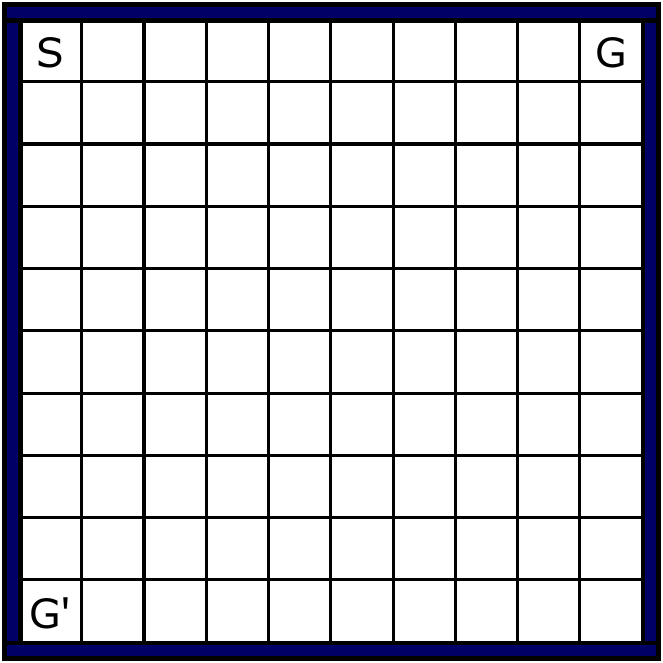}
        \caption{SG} \label{fig:simple_gridworld}
    \end{subfigure}
    \begin{subfigure}{0.14\textwidth}
        \centering
        \includegraphics[height=1.5cm]{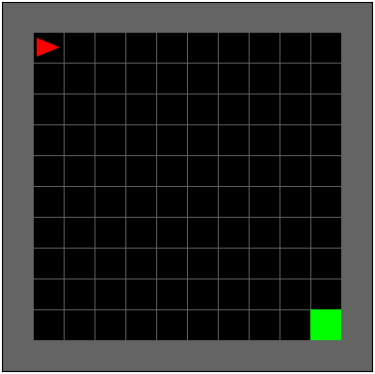}
        \caption{OneRoom} \label{fig:emptyroom}
    \end{subfigure}
    \begin{subfigure}{0.14\textwidth}
        \centering
        \includegraphics[height=1.5cm]{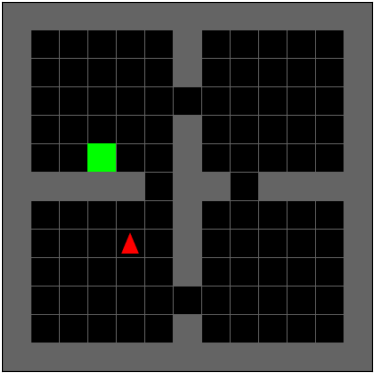}
        \caption{FourRooms} \label{fig:fourrooms}
    \end{subfigure}
    \begin{subfigure}{0.14\textwidth}
        \centering
        \includegraphics[height=1.5cm]{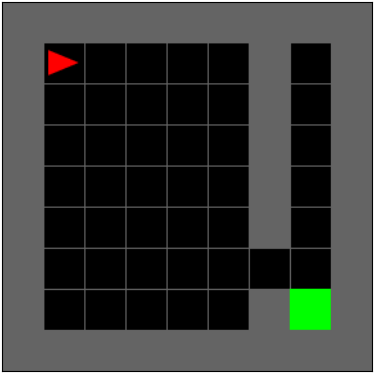}
        \caption{SCS9N1} \label{fig:simplecrossing}
    \end{subfigure}
    \par\bigskip \vspace{-0.25em}
    
    \begin{subfigure}{0.14\textwidth}
        \centering
        \includegraphics[height=1.5cm]{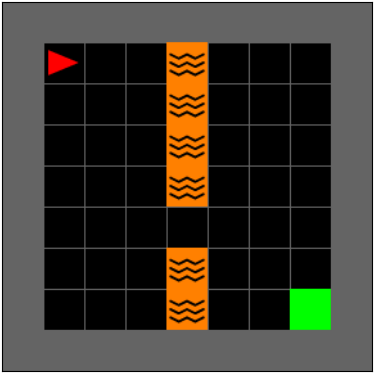}
        \caption{LCS9N1} \label{fig:lavacrossing}
    \end{subfigure}
    \begin{subfigure}{0.14\textwidth}
    \centering
        \includegraphics[height=1.5cm]{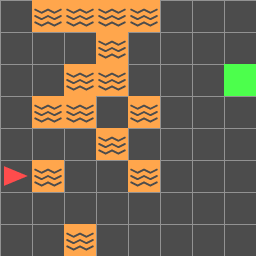}
        \caption{Source$^{0.35}$} \label{fig:rds_env_train}
    \end{subfigure}
    \begin{subfigure}{0.14\textwidth}
    \centering
        \includegraphics[height=1.5cm]{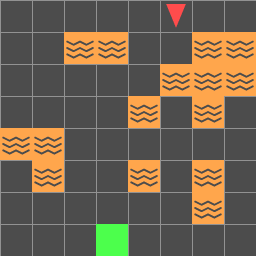}
        \caption{Target$^{0.35}$} \label{fig:rds_env_eval35}
    \end{subfigure}
    \begin{subfigure}{0.14\textwidth}
    \centering
        \includegraphics[height=1.5cm]{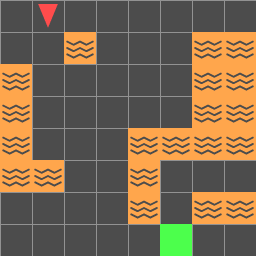}
        \caption{Target$^{0.45}$} \label{fig:rds_env_eval45}
    \end{subfigure}
    \caption{(a) The SG environment. (b-e) The OneRoom, FourRooms, SCS9N1 and LCS9N1 environments in MiniGrid. (f-h) The source task of difficulty $0.35$ and target tasks of difficulties $0.35$ and $0.45$ in the RDS environment. The difficulty parameter here controls the density of the lava cells between the agent and the goal cell, and that the target tasks are just transposed versions of the source task. With every reset of the episode, a new lava cell pattern is procedurally generated for both the source and target tasks.}
    \vspace{-0.5cm}
    \label{fig:minigrid_envs}
\end{wrapfigure}

We now perform illustrative experiments to validate the theoretical and hypothetical statements presented in Sec.\ \ref{sec:DTvsB_planning}.

\textbf{Environmental Details.} As a testbed, we use (i) the \sgenv environment (see Fig.\ \ref{fig:simple_gridworld}), and (ii) five environments from MiniGrid \citep[see Fig.\ \ref{fig:emptyroom}-\ref{fig:rds_env_eval45},][]{gym_minigrid}. We choose these environments as VE models are easy to learn in them and thus they allow for designing controlled experiments that are helpful in answering the questions of interest to this study.

In the \sgenv (SG) environment, the agent spawns in state S and has to navigate to the goal state G. At each time step, the agent receives coordinates indicating its position, and based on this selects an action that moves it to one of the four neighboring cells. The agent receives a reward of $+1$ if it reaches G and the episode terminates. In MiniGrid environments, the agent (depicted in red) has to navigate to the green goal cell, while avoiding the orange lava cells (if there are any). At each time step, the agent receives a top-down image of the grid and based on this chooses an action that moves it to one of the four neighboring cells. If the agent steps on a lava cell, the episode terminates with no reward, and if it reaches the goal cell, the episode terminates with a reward of $+1$.

\subsection{Experiments with Simplest Instantiations}
\label{sec:classic_inst_exp}

In this section, we perform experiments with the OMCP algorithm of \citet{tesauro1996line} and the Dyna-Q algorithm of \citet{sutton1990integrated, sutton1991dyna} on the SG environment to illustrate our \theorlowp presented in Sec.\ \ref{sec:classic_inst}. The implementation details of these algorithms can be found in App.\ \ref{app_sec:imp_details_of_SI}.

\begin{figure}[]
    \centering
    \begin{subfigure}{0.245\textwidth}
        \centering
        \includegraphics[height=3cm]{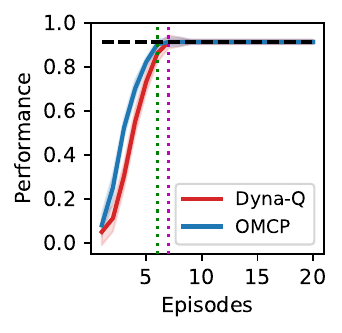}
        \vspace{-0.25cm}
        \caption{SG} \label{fig:plot_planlearn_1}
    \end{subfigure}
    \begin{subfigure}{0.245\textwidth}
        \centering
        \includegraphics[height=3cm]{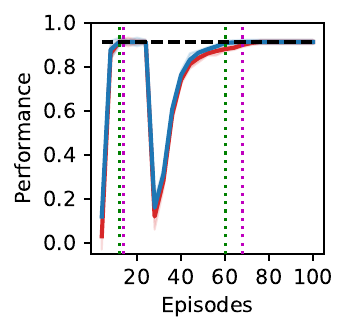}
        \vspace{-0.25cm}
        \caption{SG Adaptation} \label{fig:plot_transfer_1}
    \end{subfigure}
    \begin{subfigure}{0.245\textwidth}
        \centering
        \includegraphics[height=3cm]{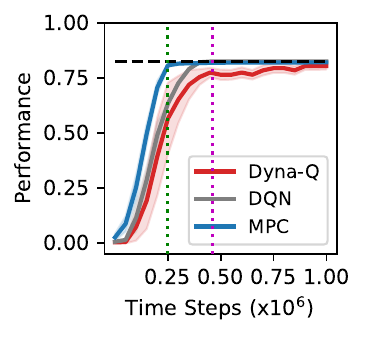}
        \vspace{-0.25cm}
        \caption{OneRoom} \label{fig:plot_emptyroom}
    \end{subfigure}
    \begin{subfigure}{0.245\textwidth}
        \centering
        \includegraphics[height=3cm]{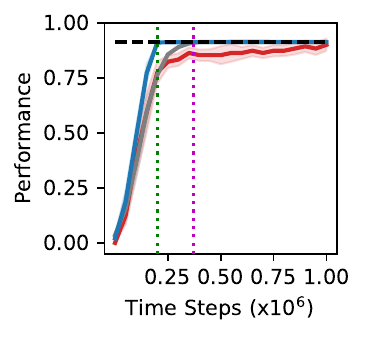}
        \vspace{-0.25cm}
        \caption{FourRooms} \label{fig:plot_fourrooms}
    \end{subfigure}
    \par\bigskip \vspace{-0.5em}
    
    \begin{subfigure}{0.245\textwidth}
        \centering
        \includegraphics[height=3cm]{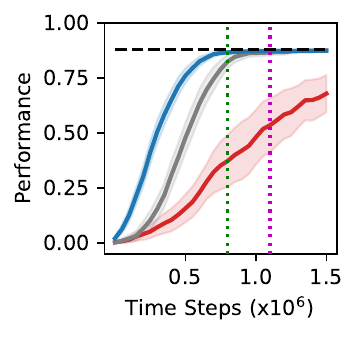}
        \vspace{-0.25cm}
        \caption{SCS9N1} \label{fig:plot_simplecrossing}
    \end{subfigure}
    \begin{subfigure}{0.245\textwidth}
        \centering
        \includegraphics[height=3cm]{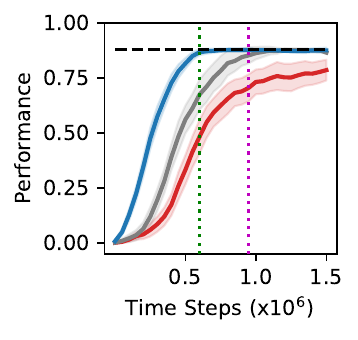}
        \vspace{-0.25cm}
        \caption{LCS9N1} \label{fig:plot_lavacrossing}
    \end{subfigure}
    \begin{subfigure}{0.245\textwidth}
        \centering
        \includegraphics[height=3cm]{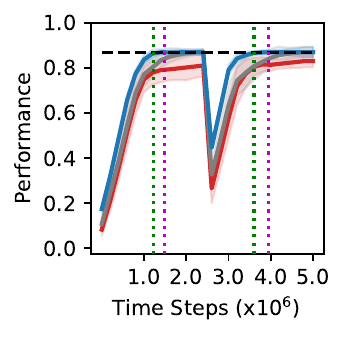}
        \vspace{-0.25cm}
        \caption{RDS Adaptation} \label{fig:plot_trans_seq}
    \end{subfigure}

    \par\bigskip \vspace{-0.5em}
    \begin{subfigure}{0.245\textwidth}
        \centering
        \includegraphics[height=3cm]{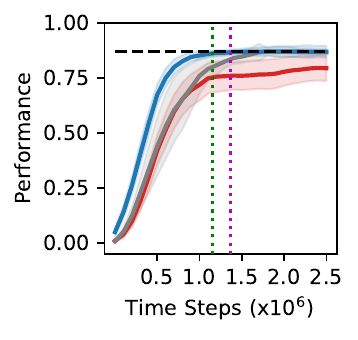}
        \vspace{-0.25cm}
        \caption{RDS Source$^{0.35}$} \label{fig:plot_trans_odist_train}
    \end{subfigure}
    \begin{subfigure}{0.245\textwidth}
        \centering
        \includegraphics[height=3cm]{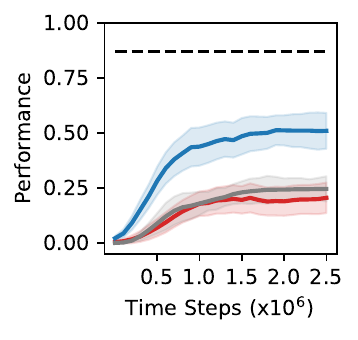}
        \vspace{-0.25cm}
        \caption{RDS Target$^{0.25}$} \label{fig:plot_trans_odist_025}
    \end{subfigure}
    \begin{subfigure}{0.245\textwidth}
        \centering
        \includegraphics[height=3cm]{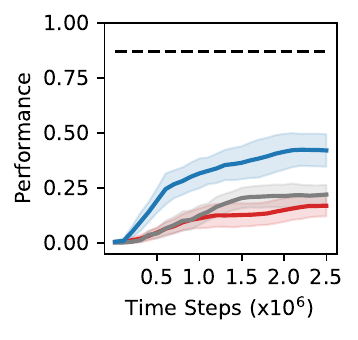}
        \vspace{-0.25cm}
        \caption{RDS Target$^{0.35}$} \label{fig:plot_trans_odist_035}
    \end{subfigure}
    \begin{subfigure}{0.245\textwidth}
        \centering
        \includegraphics[height=3cm]{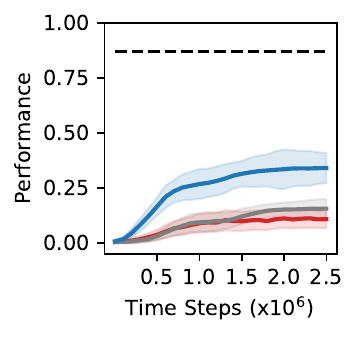}
        \vspace{-0.25cm}
        \caption{RDS Target$^{0.45}$} \label{fig:plot_trans_odist_045}
    \end{subfigure}
    \caption{(a, b) The performance of OMCP and Dyna-Q in the (a) regular RL and (b) transfer learning settings. (c-k) The performance of deep MPC, deep Dyna-Q (abbreviated as MPC and Dyna-Q) and DQN in the (c-f) regular RL and (g-k) transfer learning settings. The black dashed lines indicate the performance of the optimal policy in the corresponding environment. The green and magenta dotted lines indicate the point after which the models of the decision-time and background planning algorithms become and remain as VE models, respectively. The shaded regions are confidence intervals over (a, b) 50 and (c-k) 25 runs.}
    \label{fig:modern_algs}
\end{figure}

\textbf{Regular RL Setting.} \theor \ref{tr:tr1} states that, in the \rrl setting, the OMCP algorithm will perform on par with the Dyna-Q algorithm. For empirical illustration, we compared these algorithms on the SG environment. The empirical result in Fig.\ \ref{fig:plot_planlearn_1} shows that, as expected, the OMCP algorithm performs on par with the Dyna-Q algorithm, which illustrates \theor \ref{tr:tr1}.

\textbf{Transfer Learning Setting.} \theor \ref{tr:tr2} states that in the adaptation setting, the OMCP algorithm will perform on par with the Dyna-Q algorithm in the source task, and the same will happen in the subsequent target task. For illustration, we compared these algorithms on an adaptation version of the SG environment. In this environment, the agent is first trained and evaluated on a source task with goal G and then it is subsequently trained and evaluated on a target task with with goal G$^{'}$ (see Fig.\ \ref{fig:simple_gridworld}). In Fig.\ \ref{fig:plot_transfer_1}, we can see that, the OMCP algorithm first performs on par with the Dyna-Q algorithm in the source task and the same happens in the target task, illustrating \theor \ref{tr:tr2}.

\subsection{Experiments with Modern Instantiations}
\label{sec:modern_inst_exp}

We now perform experiments with the deep MPC and deep Dyna-Q algorithms of \citet{zhao2021consciousness} on MiniGrid environments to empirically validate our \hypolowp in Sec.\ \ref{sec:modern_inst}. In addition to these two algorithms, we also perform experiments with the model-free DQN algorithm \citep{mnih2015human} to illustrate (i) the negative effects of the usage of simulated experiences in the deep Dyna-Q algorithm and (ii) the positive effects of performing online planning in the deep MPC algorithm. We note that the only difference between deep Dyna-Q and DQN is that in the DQN algorithm the value estimator is only updated with real experiences (see line 17-20 in Alg.\ \ref{alg:alg_SI}). We also note that the only difference between deep MPC and DQN is that in the DQN algorithm there is no online planning component (see line 12 in Alg.\ \ref{alg:alg_MI}). The implementation details of these algorithms can be found in App.\ \ref{app_sec:imp_details_of_MI}.

\textbf{Regular RL Setting.} In \hypo \ref{h:h1}, we hypothesized that in the regular RL setting, the deep MPC algorithm will perform on par or better than the deep Dyna-Q algorithm as the value estimator of the deep Dyna-Q algorithm will suffer from the negative effects of the harmful simulated experiences. To validate this hypotheses, we compared these algorithms on four MiniGrid environments: OneRoom, FourRooms, SimpleCrossingS9N1 (SCS9N1) and LavaCrossingS9N1 (LCS9N1). The results, in Fig.\ \ref{fig:plot_emptyroom}-\ref{fig:plot_lavacrossing}, show that the deep MPC algorithm indeed performs on par or better than deep Dyna-Q algorithm, even after deep Dyna-Q's model becomes a VE model, which validates \hypo \ref{h:h1}.

To make sure that the simulated experiences are causing the problem, we also trained the DQN algorithm on these environments. Results in Fig.\ \ref{fig:plot_emptyroom}-\ref{fig:plot_lavacrossing}, show that the DQN algorithm achieves the same performance with the deep MPC algorithm, illustrating the negative effects of using simulated experience in the updates of the value estimator in the deep Dyna-Q algorithm.

\textbf{Transfer Learning Setting.} In Sec.\ \ref{sec:modern_inst}, we first hypothesized in \hypo \ref{h:h2} that in both the adaptation and generalization settings, a similar performance trend with the regular RL setting will hold on the source tasks. Then, we hypothesized in \hypo \ref{h:h3} that in the adaptation setting, the deep Dyna-Q algorithm will suffer more in adaptation process and thus perform worse than the deep MPC algorithm on the target tasks. Finally, we hypothesized in \hypo \ref{h:h4} that in the generalization setting, the deep MPC algorithm will improve upon its existing policy and perform better than the deep Dyna-Q algorithm on the target tasks.

In order to test the validity of \hypo \ref{h:h2} and \ref{h:h3}, which are related to the adaptation setting, we compared the deep MPC and deep Dyna-Q algorithms on an adaptation version of the RandDistShift \citep[RDS,][]{zhao2021consciousness} environment. In this environment, the agent is first trained and evaluated on source tasks with difficulty $0.35$ (see Fig.\ \ref{fig:rds_env_train}) and then it is subsequently trained and evaluated on target tasks with difficulty $0.35$ (see Fig.\ \ref{fig:rds_env_eval35}). Results in Fig.\ \ref{fig:plot_trans_seq} show that (i) the deep Dyna-Q algorithm indeed performs worse than the deep MPC algorithm on the source tasks, and (ii) it indeed suffers more in the adaptation process and performs worse than deep MPC on the target tasks, validating \hypo \ref{h:h2} and \ref{h:h3}, respectively.

In order to test the validity of \hypo \ref{h:h2} and \ref{h:h4}, which are related to the generalization setting, we compared the deep MPC and deep Dyna-Q algorithms on the original RDS environment \citep{zhao2021consciousness}. In this environment, the agent is trained and evaluated on source tasks with difficulty $0.35$ (see Fig.\ \ref{fig:rds_env_train}) and during the this process it is periodically evaluated on the target tasks with difficulties varying from $0.25$ to $0.45$ (see Fig.\ \ref{fig:rds_env_eval35}-\ref{fig:rds_env_eval45}). Results are shown in Fig.\ \ref{fig:plot_trans_odist_train}-\ref{fig:plot_trans_odist_045}. As can be seen in Fig.\ \ref{fig:plot_trans_odist_train}, the deep Dyna-Q algorithm again performs worse than the deep MPC algorithm on the source tasks, further validating \hypo \ref{h:h2}. We can also see in Fig.\ \ref{fig:plot_trans_odist_025}-\ref{fig:plot_trans_odist_045} that deep MPC indeed achieves significantly better generalization performance than deep Dyna-Q across all target tasks with varying difficulties, validating \hypo \ref{h:h4}.

To make sure that the simulated experiences are causing the problem in the deep Dyna-Q algorithm and that the online planning component is leading to a generalization performance improvement in the deep MPC algorithm, we also trained the DQN algorithm on the RDS environments. Results in Fig.\ \ref{fig:plot_trans_seq}-\ref{fig:plot_trans_odist_train}, show that the DQN algorithm achieves the same performance with the deep MPC algorithm, again illustrating the negative effects of using simulated experience in the updates of the value estimator in the deep Dyna-Q algorithm. We can also see in Fig.\ \ref{fig:plot_trans_odist_025}-\ref{fig:plot_trans_odist_045} that performing online planning significantly boosts the performance of the deep MPC algorithm compared to the DQN algorithm.

\section{Related Work}
\label{sec:related_work}

In the model-based RL literature, there have been benchmarking studies that empirically compare the performances of various decision-time and background planning algorithms on continuous control domains in the regular RL setting \citep{wang2019benchmarking} and on MiniGrid environments in specific transfer learning settings \citep{zhao2021consciousness}. However, unlike our study, none of these studies provide a unified analysis of when will one planning method perform better than the other across different settings. Also, rather than comparing the algorithms using the expected discounted return, these studies perform the comparison using the expected undiscounted return, and thus might be misleading in understanding the degree of optimality of the generated output policies.

The study of \citet{moerland2023model} reviews many of the decision-time and background planning algorithms proposed in the model-based RL literature. However, it does not focus on providing an answer to the question of how would the two prevalent planning methods in model-based RL compare against each other across different settings; in contrast, this study exactly focuses on this question.

The categorization within the decision-time planning algorithms that we provide in this study (see Sec.\ \ref{sec:background}) has connections to the recent monograph of \citet{bertsekas2021rollout} in which the recent successes of AlphaZero \citep{silver2018general}, a decision-time planning algorithm, are viewed through the lens of dynamic programming. However, we take a broader perspective and provide a categorization that encompasses both decision-time and background planning algorithms. Also, in this study, instead of assuming the availability of an exact model, we consider scenarios in which a model has to be learned by pure interaction with the environment. 

Another related study is the study of \citet{hamrick2021role} which aims for providing an understanding of how the different components of decision-time algorithms can affect their performance across different settings by studying the MuZero algorithm \citep{schrittwieser2020mastering}. Rather than just considering decision-time planning algorithms, in this study we also consider background planning algorithms and we provide a discussion on how one of the key characteristics of background planning can hinder its performance across different settings.

\section{Conclusion and Discussion}
\label{sec:conclusion}

To summarize, we performed a unified analysis of value-based decision-time and background planning methods and attempted to answer the following question: \textit{using the discounted return as the performance measure and considering value-based versions of these methods, in which settings will one planning method perform better than the other?} In our analysis, we have studied generic algorithms and provided theoretical and hypothetical statements that are generally applicable to most other instantiations of the two value-based planning methods in their corresponding classes. Overall, our findings suggest that even though value-based decision-time and background planning methods perform on par on their simplest instantiations, the modern instantiations of value-based decision-time planning methods can perform on par or better than the modern instantiations of value-based background planning methods in both the regular RL and transfer learning settings.

We note that the main purpose of this study was to contribute towards the goal of providing a general understanding of in which settings will one value-based planning method perform better than the other through studying the generic algorithms in their corresponding classes, and not to provide a benchmark that compares state-of-the-art model-based RL algorithms across various settings and domains. We also note that even though providing practical insights is not the main goal of this study, we believe that our study can guide the community in improving value-based background planning methods in potentially interesting ways. For example, a possible improvement to modern background planning algorithms could be to add a meta-level algorithm that controls the amount of simulated data in the updates of the value estimator throughout learning. Finally, we note that we were only interested in comparing the two planning methods in terms of the expected discounted return of their output policies; though not the main focus of this study, other possible interesting comparison directions include comparing the two methods in terms of their sample efficiency and real-time performance.

\bibliography{ewrl}
\bibliographystyle{plainnat}

\clearpage
\appendix
\counterwithin{figure}{section}
\counterwithin{table}{section}

\section{Proofs}
\label{app_sec:proofs}

\begin{appproposition} 
    \label{app_prop:prop}
    Let $m_{O}\in\mathcal{M}$ and $m_{D}\in\mathcal{M}$ denote the models of the of the OMCP and Dyna-Q algorithms, respectively, and let $\pi_{m_O}$ and $\pi_{m_D}$ denote their final policies generated as a result of planning with their corresponding models. Then, when $m_O$ and $m_D$ become VE models of $m^*\in\mathcal{M}$, we have $J_{m^*}^{\pi_{m_O}} = J_{m^*}^{\pi_{m_D}}$.
\end{appproposition}
\begin{proof}
    This result directly follows from the definition of VE models (see Sec.\ \ref{sec:background}). Recall that when a model $m\in \mathcal{M}$ is a VE model, the following equality holds:
    \begin{equation*}
        V_{m^*}^{\pi_{m}^*} = V_{m^*}^* \quad \forall \pi_{m}^*\in\mathbb{\Pi},
    \end{equation*}
    which implies $V_{m^*}^{\pi_{m_O}^*} = V_{m^*}^{\pi_{m_D}^*} = V_{m^*}^*$ for all $\pi_{m_O}^*, \pi_{m_D}^* \in\mathbb{\Pi}$. This in turn implies $J_{m^*}^{\pi_{m_O}} = J_{m^*}^{\pi_{m_D}}$.
\end{proof}

\section{Implementation Details}
\label{app_sec:imp_details}

In this section, we provide the implementation details of the simplest and modern instantiations of the two planning methods that were considered in this study.

\subsection{Implementation Details of the Simplest Instantiations}
\label{app_sec:imp_details_of_SI} 

As stated in the main text, for our simplest instantiation experiments, we considered the OMCP algorithm of \citet{tesauro1996line} and the Dyna-Q algorithm of \citet{sutton1990integrated, sutton1991dyna}. The joint pseudocode that shows the similarities and differences between these two algorithms is presented in Alg.\ \ref{alg:alg_SI}. The details and hyperparameters of these algorithms are presented in Table \ref{tab:SI_details}.

\begin{table}[H]
    \algfontsize
    \caption{Details and hyperparameters of the OMCP and Dyna-Q algorithms presented in Alg. \ref{alg:alg_SI}. The \textcolor{newblue}{blue} and \textcolor{red}{red} colored parts are specific to the \textcolor{newblue}{OMCP} and \textcolor{red}{Dyna-Q} algorithms, respectively.}
    \centering
    \begin{tabular}{|l|l|}
        \hline
        \textbf{Details and hyperparameters} & \textbf{Values} \\ \hline
        representation of $Q$ & tabular value function with $|\mathcal{S}|\times|\mathcal{A}|$ entries \\
        representation of $m$ & tabular model with $|\mathcal{S}|\times|\mathcal{A}|$ entries \\
        \textcolor{newblue}{number of episodes to perform to perform rollouts, $n_r$} & \textcolor{newblue}{50}  \\
        \textcolor{red}{number of iterations to perform planning, $n_p$} & \textcolor{red}{100} \\
        exploration rate, $\epsilon$ & linearly decays from $1.0$ to $0.0$ over 5 episodes \\
        learning rate & 1e-1 \\
        discount factor & 0.95 \\
        \hline
    \end{tabular}
    \label{tab:SI_details}
\end{table}

Both of these algorithms use the update rule of tabular Q-learning for updating $Q(S,A)$:
\begin{equation*}
    Q(S, A) \gets Q(S, A) + \alpha \left[ R + \gamma \max_{a'\in \mathcal{A}} Q(S', a') - Q(S, A) \right],
\end{equation*}
and a regular update rule from supervised learning for updating $m(S,A)$:
\begin{equation*}
    m(S, A) \gets S', R, \text{done}.
\end{equation*}

\subsection{Implementation Details of the Modern Instantiations}
\label{app_sec:imp_details_of_MI}

\begin{table}[]
    \footnotesize
    \centering
    \caption{The neural network architectures of the value estimators and models of the deep MPC and deep Dyna-Q algorithms.}
    \begin{tabular}{|c|c|c|}
        \hline
        \multirow{20}{5em}{Components}
         & Value estimator  & \begin{tabular}[h]{@{}c@{}}CNN:\\ (Channels: $[32 \times 64 \times 64]$\\
        Kernel Sizes: $[8 \times 3 \times 3]$\\
        Strides: $[4 \times 2 \times 2]$),\\
        Followed by MLP: $[512]$,\\ Activation Function: ReLU\end{tabular}\\
        \cline{2-3}
        & Dynamics model  & \begin{tabular}[h]{@{}c@{}}CNN:\\ (Channels: $[32 \times 64 \times 64]$\\
        Kernel Sizes: $[8 \times 3 \times 3]$\\
        Strides: $[4 \times 2 \times 2]$),\\
        Followed by Transposed CNN:\\(Channels: $[64 \times 32 \times 3]$\\
        Kernel Sizes: $[6 \times 6 \times 5]$\\
        Strides: $[1 \times 4 \times 3]$),\\ Activation Function: ReLU \end{tabular}\\
        \cline{2-3}
        & Reward model  & \begin{tabular}[h]{@{}c@{}}CNN:\\ (Channels: $[32 \times 64 \times 64]$\\
        Kernel Sizes: $[8 \times 3 \times 3]$\\
        Strides: $[4 \times 2 \times 2]$),\\
        Followed by MLP: $[512]$,\\ Activation Function: ReLU \end{tabular}\\
        \cline{2-3}
        & Termination model  & \begin{tabular}[h]{@{}c@{}}CNN:\\ (Channels: $[32 \times 64 \times 64]$\\
        Kernel Sizes: $[8 \times 3 \times 3]$\\
        Strides: $[4 \times 2 \times 2]$),\\
        Followed by MLP: $[512]$,\\ Activation Function: ReLU \end{tabular} \\
        \hline
    \end{tabular}
    \label{tab:arch1}
\end{table}

As stated in the main text, for our modern instantiation experiments, we considered the deep MPC and deep Dyna-Q algorithms in \citet{zhao2021consciousness}, which are referred to as ``MPC (UP)'' and ``Dyna'' in \citet{zhao2021consciousness}. The joint pseudocode that shows the similarities and differences between these two algorithms is presented in Alg.\ \ref{alg:alg_MI}. The details and hyperparameters of these algorithms are presented in Table \ref{tab:MI1_details}.

\begin{table}[H]
    \algfontsize
    \caption{Details and hyperparameters of the deep MPC and deep Dyna-Q algorithms presented in Alg. \ref{alg:alg_MI}. The \textcolor{newblue}{blue} and \textcolor{red}{red} colored parts are specific to the \textcolor{newblue}{deep MPC} and \textcolor{red}{deep Dyna-Q} algorithms, respectively.
    }
    \centering
    \begin{tabular}{|l|l|}
        \hline
        \textbf{Details and hyperparameters} & \textbf{Values} \\ \hline
        representation of $Q_\theta$ & CNN with an architecture as in Table \ref{tab:arch1} \\
        representation of $m_\omega$ & CNN with an architecture as in Table \ref{tab:arch1}  \\
        size of $\mathcal{D}$ & 1e6 \\
        \textcolor{red}{size of $\mathcal{D}_i$} & \textcolor{red}{1e6} \\
        replay buffer size to start updates, $N$ & 5e4 \\
        \textcolor{newblue}{number of steps to perform search, $n_{s}$} & \textcolor{newblue}{5} \\
        \textcolor{red}{number of samples to sample from $\mathcal{D}_i$, $n_{ib}$} & \textcolor{red}{128} \\
        number of samples to sample from $\mathcal{D}$, $n_{b}$ & 128 \\
        \textcolor{newblue}{search heuristic, $h$} & \textcolor{newblue}{best-first search} \\
        exploration rate, $\epsilon$ & linearly decays from 1.0 to 0.0 over 1M time steps \\
        learning rate & 1e-4 \\
        optimizer type & Adam \\
        discount factor & 0.95 \\
        \hline
    \end{tabular}
    \label{tab:MI1_details}
\end{table}

In both of these algorithms, the parameters of the value estimator $Q_\theta$ are updated towards minimizing the TD loss $\mathcal{L}_{\text{TD}}(\theta)$, which is calculated in a similar fashion to DQN \citep{mnih2015human}; and the parameters of the model $m_\omega$ are updated towards minimizing the dynamics loss $\mathcal{L}_{\text{dyn}}(\omega)$, reward loss $\mathcal{L}_{\text{r}}(\omega)$ and termination loss $\mathcal{L}_{\text{t}}(\omega)$. Specifically, $\mathcal{L}_{\text{dyn}}(\omega)$ is the $\ell_2$ distance between the between the predicted next state $\hat{S}'$ and the true next state $S'$, $\mathcal{L}_{\text{r}}(\omega)$ is the $\ell_2$ distance between the between the predicted reward $\hat{R}$ and the true reward $R$, and $\mathcal{L}_{\text{t}}(\omega)$ is the binary cross-entropy loss between the imagined termination and the true termination. For more on the training details of the value estimators and models of the two algorithms, we refer the reader to \citet{zhao2021consciousness} and its publicly available code\footnote{See \url{https://github.com/mila-iqia/Conscious-Planning}.}.

In our modern instantiation experiments, we also considered the DQN algorithm \citep{mnih2015human}. This algorithm is just a version of the deep Dyna-Q algorithm in which real experiences are used in place of the simulated experiences for updating the value estimator. Its details and hyperparameters are the same with the deep Dyna-Q algorithm.

Additionally, we note that in our adaptation experiments, we reinitialized the replay buffers of all algorithms after the tasks switched from the source tasks to the target tasks.

\end{document}